# Copula-Linked Parallel ICA: A Method for Coupling Structural and Functional MRI brain Networks

Oktay Agcaoglu[1, ∗], Rogers F. Silva[1], Deniz Alacam[1], Sergey Plis[2], Tulay Adali[3], Vince Calhoun[1] , and for the Alzheimer's Disease Neuroimaging Initiative[*]

[1] Tri-institutional Center for Translational Research in Neuroimaging & Data Science (TReNDS), Georgia State, Georgia Tech, and Emory, Atlanta, GA, USA, 30303
[2] Georgia State University, Computer Science, Atlanta, GA, USA, 30303
[3] Department of Computer Science and Electrical Engineering, University of Maryland, Baltimore County, Maryland, USA, 21250
∗: Corresponding author: oagcaoglu@gsu.edu

*Abstract*— Different brain imaging modalities each offer unique insights into the complexities of brain function and structure. When these modalities are combined, they can significantly enhance our understanding of neural mechanisms. Prior multimodal studies fusing functional magnetic resonance imaging (fMRI) and structural MRI (sMRI) have shown the advantages of this approach. Typically, because sMRI data lacks temporal dimension, existing fusion methods focus on compressing all the temporal information in fMRI into a single summary measure. This can be seen in methods like fractional amplitude of low-frequency fluctuations (fALFF), regional homogeneity (ReHo), and functional network connectivity (FNC). However, these approaches sacrifice the rich temporal dynamics captured in the original four-dimensional fMRI data. Building on the observation from prior research that covarying networks identified via sMRI show some correspondence to those observed in resting-state fMRI networks, we have developed a novel fusion method by combining the strength of deep learning frameworks, copulas and independent component analysis (ICA), named copula linked parallel ICA (CLiP-ICA). This method not only estimates independent sources and an unmixing matrix for each imaging modality but also links the spatial sources of fMRI and sMRI using a novel model that incorporates copulas to build our fusion approach. This linkage allows for a more sophisticated and flexible integration of temporal and spatial information compared to the traditional methods, by allowing for varying yet similar structural-functional spatial relationships. We evaluated the efficacy of CLiP-ICA through rigorous testing in both simulated and real-world scenarios, utilizing data from the Alzheimer's Disease Neuroimaging Initiative (ADNI), including both fMRI and sMRI datasets. Our findings demonstrated that CLiP-ICA could effectively capture both strongly and weakly linked sMRI and fMRI networks, including the cerebellum, sensorimotor, visual, cognitive control and the default mode networks. The CLiP-ICA revealed a greater number of meaningful components and fewer artifact components (50 resting state networks and 45 intrinsic structural networks out of 75), helping to address the long-standing optimum model order problem in ICA. Additionally, CLiP-ICA detected complex FNC patterns across different stages of cognitive decline, with CN subjects generally exhibiting higher connectivity within the sensorimotor and visual networks compared to those with AD. Interestingly, a transient increase in connectivity was observed in the mild cognitive impairment (MCI) stage, suggesting potential compensatory mechanisms. Additionally, the method introduced a new metric, structural network covariation (SNC), to measure structural atrophy on the functionally coupled networks. Overall, CLiP-ICA offers a robust approach for linking multimodal data with mismatched dimensionality, revealing meaningful neural relationships that may remain hidden with traditional methods.

## I. INTRODUCTION

Independent component analysis (ICA) is a widely used blind source separation method for studying brain imaging data from various modalities including functional magnetic resonance imaging (fMRI), structural MRI (sMRI), magnetoencephalography (MEG), positron emission tomography (PET), diffusion MRI (dMRI) and electroencephalography (EEG) (Wu et al., 2010; Agcaoglu, Silva, et al., 2022; Cetin et al., 2016; Rashid et al., 2019; Fang et al., 2021; Treacher et al., 2021; Calhoun, & Adalı, 2012). ICA decomposes brain imaging data into maximally independent networks by assuming that the observed data is a mixture of independent sources. It aims to identify these sources by maximizing their statistical independence based on known or assumed probability distributions. The independence assumption implies that the joint probability distribution of the sources is the product of their individual distributions. Our research primarily utilizes the linear mixing model, with the goal of deriving an optimal linear unmixing matrix that effectively separates the mixed signals into maximally statistically independent components (Hyvarinen et al., 2000).

Several extensions of ICA have been created for multimodal fusion of brain imaging data, including Joint ICA (jICA) (Calhoun et al., 2009), Parallel ICA (Liu et al., 2007), Parallel multilink joint ICA (Khalilullah et al., 2023; Khalilullah et al., 2024) canonical correlation analysis (CCA) (Sui et al., 2018; Correa et al., 2008), independent vector analysis (IVA) (Adali et al., 2014; Kim et al., 2006; Li et al., 2023), multi-view ICA (Richard et al., 2020), Bayesian linked tensor ICA (LICA) (Groves et al., 2011), and multiset independent subspace analysis (Silva et al., 2021). However, existing methods may still fail to capture the full interplay between sources in different modalities. For example, methods like CCA focus on uncorrelated sources, ignoring higher order statistical information. jICA assumes modalities have the same mixing matrices, CCA and IVA also often assume specific distributional forms for connected sources in multiple methods. Additionally, these methods do not fully consider the complete information in the case of mismatched dimensionality (e.g., 3D vs 4D), often pre-reducing the 4D fMRI data to 3D features, such as fractional amplitude of low frequency fluctuations (fALFF) or regional homogeneity, and thus do not consider the temporal information in the data fusion. Even in some deep learning approaches, (Fedorov et al., 2024) used reduced fMRI features, fALFF, while fusing structural and functional data.

A recent method called parallel group ICA + ICA (Qi et al., 2019) focuses on 4D fMRI data by maximizing the correlation between the loading parameters of sMRI and the temporal variation in fMRI loading parameters. However, this approach does not leverage the similarity in spatial patterns between fMRI and sMRI (Luo, Sui, Abrol, Chen, et al., 2020). Semi-blind or spatially constrained approaches (Du et al., 2013; Du et al., 2020; Calhoun et al., 2005) can help with this i.e., using a common template in a spatially constrained ICA framework, allowing the fusion of sMRI with complete 4D fMRI data, but have not fully been introduced within a fusion setting (Kotoski et al., 2024).

Expanding on these concepts, and motivated by prior work that showed covarying sMRI networks resemble resting fMRI networks (Luo, Sui, Abrol, Chen, et al., 2020), we recently introduce a copula-based joint ICA model (Agcaoglu, Silva, Alacam, et al., 2024), which proves effective in identifying true sources under noisy conditions by leveraging connections between components from different modalities. Recently, we proposed a method tailored for scenarios where loading parameters are distinct, and sources exhibit topological similarity of varying degrees linked via a copula model (Agcaoglu, Silva, Alaçam, et al., 2024). This model is termed copula-linked Parallel ICA (CLiP-ICA). CLiP-ICA capitalizes relationships across different modalities, estimating linked components with higher model order and simultaneously determining functional network connectivity (FNC).

In this journal extension, we applied our original CLiP-ICA model with additional analyses to ensure robustness and enhance clinical relevance through a more detailed examination. These analyses include a replicability assessment, in which we ran CLiP-ICA on the dataset ten times, using the ICASSO framework (Himberg et al., 2004) to confirm component stability across runs and selecting the most stable components for further analysis. This journal version also places a greater emphasis on selecting clinically relevant resting-state components, resulting in a refined classification of artifact and non-artifact brain networks for both sMRI and fMRI, as well as differences in the number of meaningful components identified. Additionally, we refined the simulations to better demonstrate a range of dependency conditions.

We also introduced a novel metric, Structural Network Covariation (SNC), designed to quantify structural atrophy within functionally coupled networks, providing new insights into structural-functional relationships, which is particularly relevant to aging and Alzheimer's disease. To improve interpretability, we expanded the statistical comparisons across dementia subgroups, offering a deeper understanding of structural-functional coupling differences associated with neurodegeneration.

These developments establish CLiP-ICA as a versatile tool for robust multimodal analysis, CLiP-ICA demonstrated improved estimation of linked sMRI/fMRI components while effectively capturing unique, modality-specific (unlinked) components. validated here using real-world data from the Alzheimer's Disease Neuroimaging Initiative (ADNI) study (Jack et al., 2008). This journal extension expands the method's utility beyond our initial work, enhancing its clinical applicability for studying neurodegeneration and aging.

## II. METHODS

In this section, we begin by introducing the CLiP-ICA method (Agcaoglu, Silva, Alaçam, et al., 2024), providing a detailed explanation of its theoretical foundation and the formulation of the cost function. We then discuss the practical applications of the method, highlighting its relevance to multimodal neuroimaging data analysis.

Following the theoretical exposition, we describe a simulation designed to evaluate the performance of CLiP-ICA, demonstrating its effectiveness in controlled settings. Finally, we apply the method to a real-world dataset, utilizing sMRI and fMRI data from the ADNI dataset to showcase its utility in extracting meaningful brain networks.

### A. Copula Linked Parallel ICA for multimodal data

The primary goal of this method is to fuse different modalities, which can have different temporal dimensions, by identifying two unmixing matrices that maximize the likelihood of a 2D joint distribution. This distribution is formed by combining the 1D marginal distribution of each imaging modality with a copula that establishes a connection between them (Sklar, 1959). The marginal distribution can vary across different imaging modalities, and the copula can take various forms, such as a Gaussian copula or a Farlie-Gumbel-Morgenstern (FGM) copula (Sklar, 1959; Ma et al., 2007). In this study, we implemented the Gaussian copula for its simplicity and its ease of extension to high dimensions, which enhances computational speed by eliminating the need to loop through each pair. The Gaussian copula is particularly advantageous because it can effectively capture the dependency structure between the marginal distributions without being overly complex, thus making it computationally efficient for large datasets and more easily interpreted.

We selected the logistic distribution as the marginal distribution based on previous ICA studies in brain imaging, as it has shown robust and replicable results across different imaging modalities and is a good match to the underlying source distributions (Cetin et al., 2016; Rashid et al., 2019; Agcaoglu, Silva, Alacam, et al., 2024; Agcaoglu et al., 2016; Allen et al., 2018; Calhoun, & Adali, 2012; McKeown et al., 1998; Esposito et al., 2002; McKeown et al., 2003; Lange et al., 1999). The logistic distribution is known for its heavier tails compared to the normal distribution, which can provide a better fit for the data characteristics typically observed in neuroimaging. This selection helps in capturing the intrinsic variability and dependencies within the brain imaging data more accurately.

By integrating the Gaussian copula with logistic marginals, our method not only leverages the theoretical strengths of copula-based models but also ensures practical applicability and reliability in neuroimaging contexts. This dual advantage makes our approach a powerful tool for uncovering the complex interdependencies in brain imaging data, ultimately contributing to a deeper understanding of brain function and structure. Our methods can also be extended to other types of data beyond neuroimaging, where the relationship between different perspectives or variables can be modeled using copulas. The flexibility of the copula framework allows for various extensions and adaptations, making it a versatile choice for multivariate data analysis in different scientific fields.

The CLiP-ICA method begins by applying principal component analysis (PCA) separately to each modality, thereby reducing the dimensionality to the final model order. Subsequently, the method estimates optimized unmixing matrices.

During the estimation phase, the batches contain data from corresponding voxels in both sMRI and fMRI modalities. To facilitate this correspondences, we adjusted the resolution of the sMRI images to match that of the fMRI images post-PCA. During the initial epochs of the estimation, we compute the cross-correlation between the sMRI and fMRI components. The columns of the weight matrices are permuted to align the best matching components. Such alignment is vital, especially in higher-order models, where the incentive for independence can overshadow the reward for copula correlation. Additionally, to account for the sign ambiguity, the corresponding column and row is multiplied by -1 where necessary to ensure positive spatial correlation between modality pairs. This helps the model to converge to the global rather than a local minimum during estimation..

Detailed steps for CLiP-ICA:

1. Dimensionality Reduction with PCA: Each modality (sMRI and fMRI) undergoes PCA independently, reducing the dimensionality to the desired model order. This step simplifies the data while preserving essential features.

2. Resolution Matching: The sMRI image resolution is adjusted to match the fMRI image resolution after PCA. This ensures that data from both modalities are comparable and can be integrated effectively.

3. Estimation and Optimization: The estimation process involves iterating through batches containing data from corresponding voxels from both sMRI and fMRI. The goal is to find the optimized unmixing matrices that maximize the joint likelihood.

4. Component Alignment: During the initial estimation epochs, we calculate the cross-correlation between the sMRI and fMRI components. Then we permute the columns of the weight matrices to match the best corresponding components. Aligning the components is a critical step, particularly in higher-order models. It ensures that the incentive for independence in the components does not surpass the reward for maintaining the copula correlation. Proper alignment prevents the model from settling into a local minimum, guiding it towards the global minimum during estimation. As mentioned earlier, the columns are also multiplied with -1 where they have negative correlation to ensure positive correlation between pairs, because in gaussian copula negative and positive correlation has the same loss.

5. Evaluation: The alignment and optimization steps are continuously evaluated to ensure that the model is converging correctly. This involves monitoring the loss function and making necessary adjustments to the estimation process.

The combination of these steps ensures that the CLiP-ICA method effectively integrates information from both sMRI and fMRI modalities, leveraging the strengths of each to achieve a robust and accurate model. A block diagram illustrating the CLiP-ICA method is presented in Figure 1, and the derivation of the loss function is detailed below.

The derivation of the loss function of the CLiP-ICA can be described as follows. Let $\mathbf{X_1}$ and $\mathbf{X_2}$ be the dimension-reduced data of modality 1 and 2, each with a size of $c \times v$, and let $\mathbf{X}$ be the concatenated data of size $2c \times v$. Similarly, let $\mathbf{Y_1}$ and $\mathbf{Y_2}$ be the sources of each modality, and $\mathbf{Y}$ be the concatenation of $\mathbf{Y_1}$ and $\mathbf{Y_2}$ with a size of $2c \times v$. Let $\mathbf{W_1}$ and $\mathbf{W_2}$ be the $c \times c$ mixing matrices of each modality, respectively; such that $\mathbf{Y_1} = \mathbf{W_1} \cdot \mathbf{X_1}$ and $\mathbf{Y_2} = \mathbf{W_2} \cdot \mathbf{X_2}$ through matrix multiplication, which can be written as:.

$$\mathbf{Y} = \begin{bmatrix} \mathbf{Y_1} \\ \mathbf{Y_2} \end{bmatrix} = \begin{bmatrix} \mathbf{W_1} & 0 \\ 0 & \mathbf{W_2} \end{bmatrix} \cdot \begin{bmatrix} \mathbf{X_1} \\ \mathbf{X_2} \end{bmatrix} \text{ so that, } \mathbf{Y} = \mathbf{W} \cdot \mathbf{X}.$$

For simplicity, let $x_1$, $x_2$, $x$, $y_1$, $y_2$ and $y$ be column vectors representing samples taken from the corresponding data columns. Applying the probability transformation formula, the probability of observing $x$ for a given $W$ can be written as:

$$p_X(x) = p_Y(y) \cdot |\det W| = p_Y(y) \cdot |\det(\mathbf{W_1} \cdot \mathbf{W_2})|,$$

where $p_X(x)$ and $p_Y(y)$ are probability distribution functions. Applying the assumption that sources are independent within each modality, this can be written as:

$$p_X(x) = \prod_{i=1}^{c} p_Y(y_i, y_{i+c}) \cdot |\det W|.$$

Utilizing Sklar's theorem (Sklar, 1959), the expression can be written using a copula $c_u$ as:

$$p_X(x) = \prod_{i=1}^{c} p_y(y_i) \cdot p_y(y_{i+c}) \cdot c_u(u_i, u_{i+c}) |\det W|.$$

For the Gaussian copula:

$$c_{\mathbf{R}_i}^{\text{Gcop}}(u_i, u_{i+c}) = \frac{1}{\sqrt{\det \mathbf{R}_i}} \cdot \exp\left(-\frac{1}{2} \begin{bmatrix} \Phi^{-1}(u_i) \\ \Phi^{-1}(u_{i+c}) \end{bmatrix}^T [\mathbf{R}_i^{-1} - \mathbf{I}] \begin{bmatrix} \Phi^{-1}(u_i) \\ \Phi^{-1}(u_{i+c}) \end{bmatrix}\right),$$

where $\mathbf{I}$ is the identity matrix, and $\mathbf{R}_i$ is the correlation matrix where $\sigma_{i,i+c}$ describing the dependencies between $y_i$ and $y_{i+c}$. The variable $u_i$ refers to a variable taken from the uniform distribution after passing the sources to their cumulative distribution function (cdf), $F_{Y_i}(y)$ and $\Phi^{-1}$ is the inverse cdf of 1D Gaussian distribution with zero mean and unit variance

$$\mathbf{R}_i = \begin{bmatrix} 1 & \sigma_{i,i+c} \\ \sigma_{i,i+c} & 1 \end{bmatrix} \text{ where } |\sigma_{i,i+c}| \leq 1 \text{ and } u_i = F_{Y_i}(y_i).$$

Then the loss function can easily be optimized by minimizing negative log-likelihood, $-\log(p_X(x))$. The same equation can also be expressed in vector form with an $\mathbf{R}$ matrix of size $2c \times 2c$, all diagonal elements are $1$, indices of $(i, i + c)$ and $(i + c, i)$ are set to dependencies, and all others are set to $0$. The optimization process for the CLiP-ICA's loss function was executed in Python using the PyTorch framework (Paszke et al., 2019), with the Adam optimizer (Kingma et al., 2015).

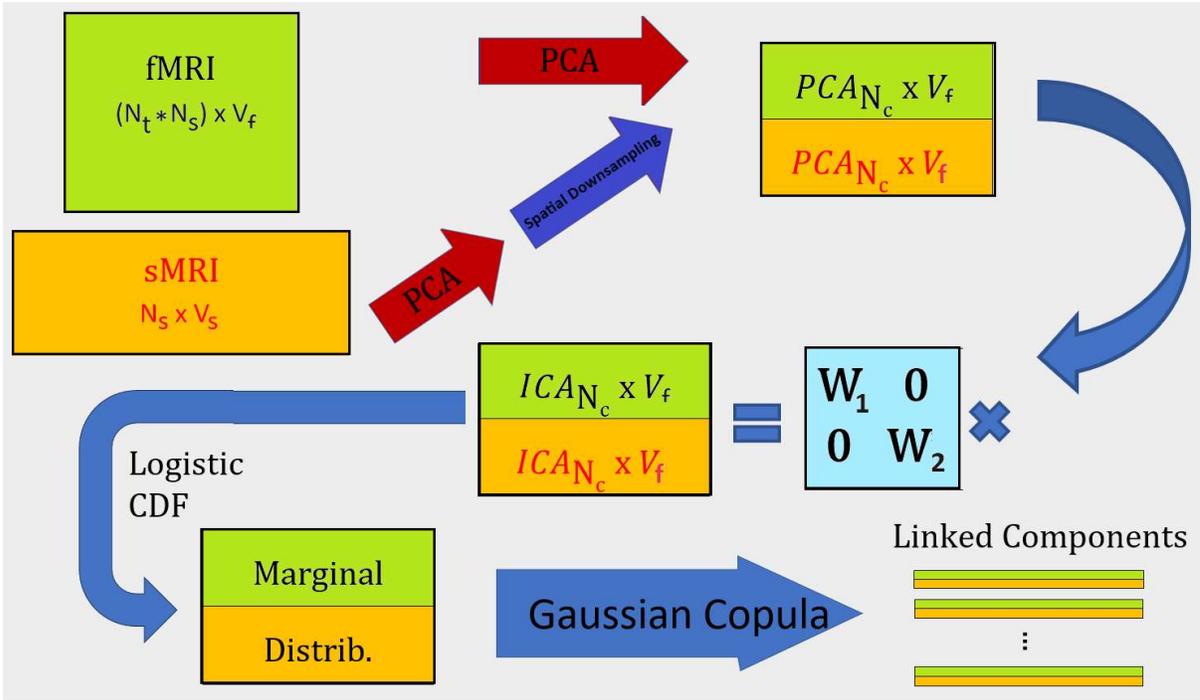

Figure 1: Block diagram of the CLiP-ICA method. Two separate PCA transformations have been applied to reduce the observed dimension (temporal/subjects) to the desired model order. In our example, the sMRI data was downsampled to match the fMRI spatial resolution to further reduce computational complexity (note, one could instead opt to upsample to preserve spatial resolution). Next, CLiP-ICA estimates the linked components

*B. Simulation*

We first tested the performance of the CLiP-ICA in a controlled environment, using a simple simulation. To do so, we generated four spatially independent components using the simulation toolbox (simTB) (Erhardt et al., 2012) (http://trendscenter.org/software). Subsequently, employing custom MATLAB scripts, we generated a second set of four components linked to the first set with Pearson correlations of 0.94, 0.94, 0.91, and 0.02. Importantly, we maintained independence between the components and preserved the blob-like structure in each component. Each component was designed with a logistic distribution with a zero mean and a scale parameter of 1. Following this, we created a mixing matrix of size 300 by 4 and generated the observed fMRI data, corresponding to 10 subjects, each with 30 volumes. Next, we generated another mixing matrix of size 10 by 4 and obtained the observed sMRI data, corresponding to 10 subjects. Finally, we attempted to recover the sources using the CLiP-ICA method. Initially, we applied two separate PCA steps to reduce the dimension to four, then z-scored the principle components. Then applied CLiP-ICA, with the Gaussian copula and dependency parameter set to 0.90 for all four components.

*C. Real Data from ADNI sMRI and fMRI*

We tested our new method using fMRI and sMRI data from the ADNI dataset (Jack et al., 2008). Both modality datasets were already preprocessed in a previous study (Abrol et al., 2020). The fMRI data was preprocessed through a standard preprocessing pipeline including motion correction, spatial normalization, and spatial smoothing (Abrol et al., 2020). We run the sMRI data through a unified segmentation pipeline, resulting in gray matter (GM) probability density maps; the T1 images were then warped to MNI space, modulated to preserve volume changes, and smoothed using a 3D Gaussian kernel to 6 mm full width at half maximum (FWHM)(Abrol et al., 2020). Our dataset included a total of 864 people. Out of these, 83 were diagnosed with Alzheimer's disease (AD), 384 had normal cognitive abilities (CN), 105 showed early signs of mild cognitive impairment (EMCI), 188 were identified with mild cognitive impairment (MCI), 57 participants were in the late stage of mild cognitive impairment (LMCI), and 47 participants had significant memory concerns (SMC).

We applied our method to the ADNI dataset. For our analysis, we selected an ICA model with an order of 75 components, which has been shown to capture modular FNC (Allen et al., 2011; Agcaoglu et al., 2015). To eliminate the non-brain regions, we generated a common mask for the fMRI and sMRI data. The fMRI images were masked and demeaned so that each volume has zero mean. The fMRI images were underwent a two-step PCA process: first, a subject-specific PCA reduced the dimensions to 100, and then principle components were z-scored, followed by a group PCA to further reduce to 75 dimensions.

The sMRI images were initially concatenated along the subject dimension and were masked using a group sMRI mask and demeaned. Then PCA was applied to reduce to 75 dimensions. Afterward, the PCA-processed sMRI images were down-sampled to match the spatial resolution of the fMRI images. Then the common mask was used to retain same voxels with fMRI. We used Group ICA Of fMRI Toolbox (GIFT) (https://github.com/trendscenter/gift) for the PCA and AFNI's (https://afni.nimh.nih.gov/) 3dresample function to reduce the spatial resolution.

We set the Gaussian copula linkage parameters (sMRI/fMRI similarity) to span from 0.95 to 0.5, equally spaced for 75 networks (note that the networks naturally align with these linkages during the optimization step). Along with this, the copula covariance matrix was defined as 150 by 150, outlining the linkage. To ensure stability, we repeated the analysis 10 times, every time starting at random initial points and checking for the repeatability of components using ICASSO framework (Himberg et al., 2004). Then the run that provided the most stable components that are at the centroids of all runs has been selected for further analysis. After generating the aggregated components, we calibrated the components sign based on the skewness of the distribution.

Next, we performed back-reconstruction on sMRI and fMRI data using the aggregated components with a spatially constrained ICA algorithm called multi-objective optimization ICA with reference (MOO-ICAR) (Du et al., 2013; Du et al., 2020) and generated subjects time-courses and subject-specific spatial maps; and sMRI component loadings. Finally, we selected resting state networks (RSN) for functional data and intrinsic structural networks (ISN) for structural data; using a combination of correlation with previous templates (Du et al., 2020; Allen et al., 2011; Agcaoglu et al., 2019; Agcaoglu, Muetzel, et al., 2022; Iraji et al., 2023), peak location of the components, and frequency spectrum of the component time-courses.

Following back-reconstruction, we calculated FNC matrices for each subject. To do so, subject component time-courses were detrended, band-pass filtered to 0.01-0.15 Hz, and de-spiked. Finally, FNCs were calculated as Pearson correlation coefficients. Components are reordered in each subdomain to increase modularity visually. Finally, we calculated group differences on the FNC and loading parameters, using a two-sample t-test and corrected for multiple comparison using a false discovery rate (FDR, p<0.05).

III. RESULTS

In this section, we first present the simulation results, comparing the components extracted by CLiP-ICA to the ground truth components. This comparison allows us to assess the accuracy and robustness of the method in identifying independent sources coupled to varying degrees. Next, we show results from the application of CLiP-ICA to a real-world dataset, jointly estimating linked sMRI and fMRI components and the ensuing the FNC matrices, one of the key benefits of incorporating the full temporal information of fMRI into the model. We also highlight group differences observed in the loading parameters (structural) as well as the FNCs. In addition, we inspect the structural network covariations (SNC) that reveal the relationships across different modalities. Furthermore, we provide a comparative analysis of cross-modality similarity between the results obtained from CLiP-ICA and those from separate ICA runs, demonstrating the added value of our multimodal fusion approach in uncovering more integrated and biologically meaningful insights.

*A. Simulation*

Figure 2 displays the ground truth components that are generated using simTB. The simulation includes 2 modalities with 5 sources. Among these sources, 4 are linked with varying dependencies (0.96, 0.76, 0.64 and 0.31), while 1 is unlinked (0.03). The dependency values for the method were set to 0.9 for all the components pairs to test whether CLiP-ICA could accurately identify sources when the provided dependency values did not match the ground truth-both in higher and lower cases. Figure 3 shows the results: CLiP-ICA successfully retrieved all five sources. This simulation demonstrates the method's capability to accurately identify underlying sources across different modalities, highlighting the flexibility of the approach. Notably, CLiP-ICA successfully separated the components even when the set dependency value (0.9) was above or below the true cross-modal dependencies, such as in the case of a very low true dependency (0.03). This suggests that the method adapts to the data without imposing artificially high correlations when they are not present.

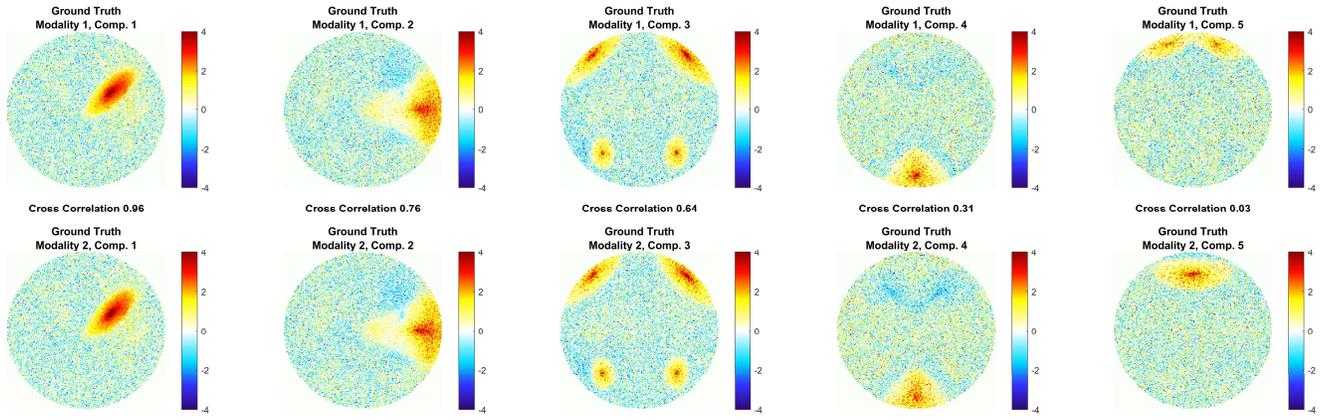

Figure 2: The ground truth components included in the simulation; the design includes a total of five components in each modality, four of these components are linked between modalities having cross-correlations (0.96, 0.76, 0.64, and 0.31) and one of them is unlinked with a cross-modality correlation of 0.03.

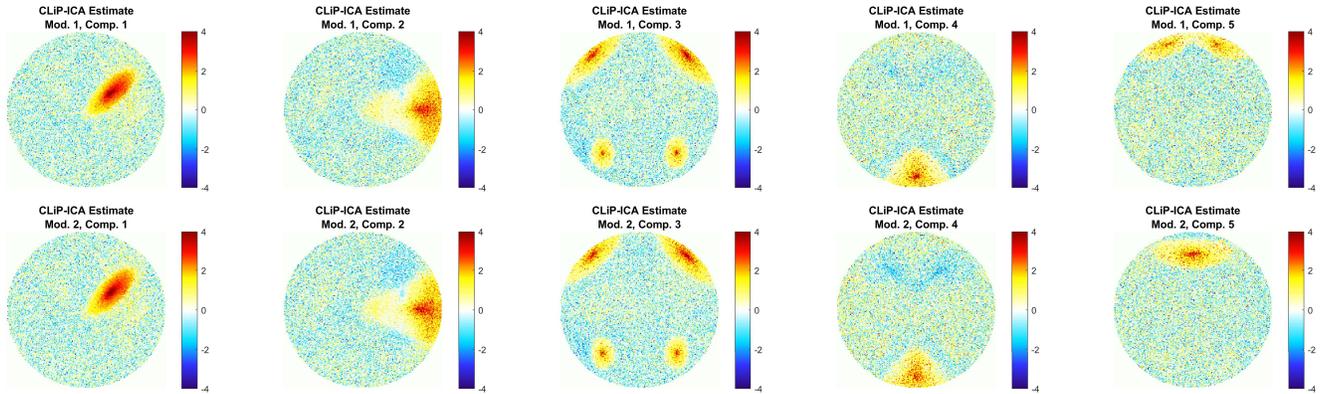

Figure 3: The CLiP-ICA estimated multimodal component pairs are displayed, demonstrating that the method successfully retrieved the original components in both linked cases with varying degrees of dependency and in the unlinked case. The simulation results illustrate the flexibility of the method, even when the assigned dependencies do not match the ground truth cross-modality dependencies.

*B. Real Data from ADNI sMRI and fMRI*

In the ADNI data, we first analyzed the degree of linkage between the aggregated components of sMRI and fMRI by calculating the spatial Pearson cross-correlation. The resulting values are displayed in Figure 4 as well as the CLiP-ICA assigned copula dependency values. For comparison, we also performed two separate group ICA analyses on the fMRI and sMRI datasets, respectively, with the same model order of 75. Post-hoc, we matched these components based on their spatial similarity, using Pearson correlation, and sorted them from highest to lowest correlation, also shown in Figure 4. The correlation values obtained from CLiP-ICA fell between the assigned copula dependency values and the cross-correlation values from the separate ICA runs. This demonstrates the influence of our proposed method on component estimation and highlights its ability to effectively link modalities.

We also assessed the stability of the estimated CLiP-ICA using the ICASSO framework (Himberg et al., 2004), and observed that the stability and reproducibility of the components were sufficient. Please refer to the Supplementary

Material for the ICASSO results, and also for the results of the estimated components by running the two separate group ICA analyses.

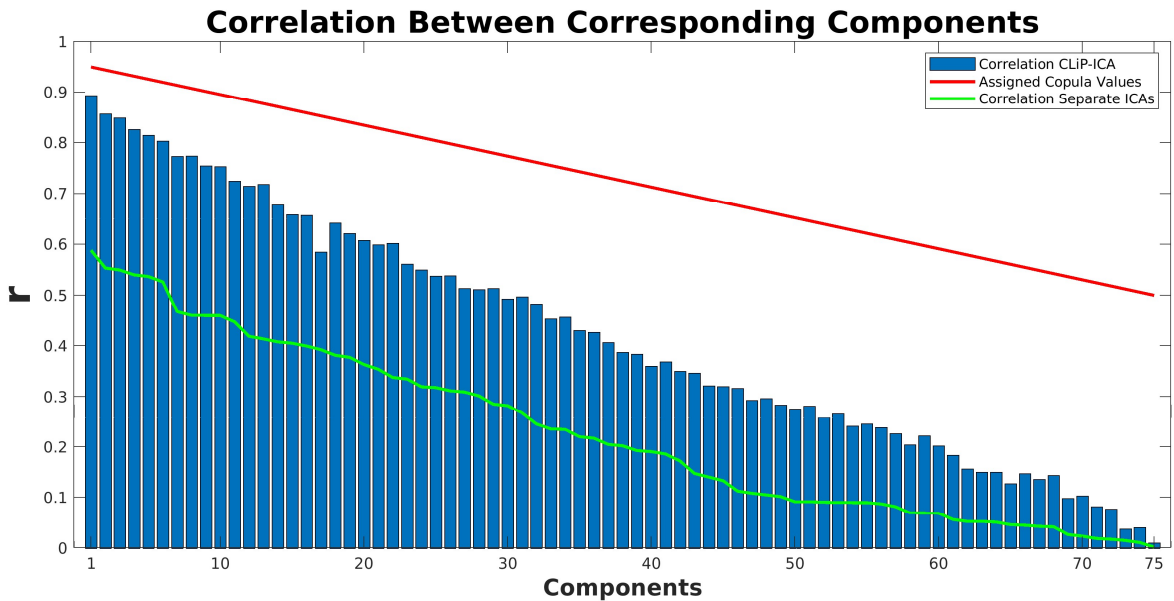

Figure 4: Assigned copula values (red), Pearson correlation graph of estimated fMRI and sMRI sources (blue), the correlation resulting from running two separate ICA with post-hoc matching (green) are shown. Some components show a strong linkage with 0.89 correlation while some have no linkage between modalities. CLiP-ICA correlation values are much higher than the separate ICA runs, which shows the impact of the method on the results.

Based on the CLiP-ICA results, we identified 50 out of 75 components as resting-state networks (RSNs) in the fMRI data and 45 out of 75 intrinsic structural networks (ISNs) from the sMRI data. First, we observed that components up to number 59 were linked between sMRI and fMRI, while those beyond this point showed a low correlation. Visual analysis further confirmed that components after 59 were effectively unlinked, indicating these are modality-unique sources between sMRI and fMRI. Consequently, we selected RSN and ISN components 60 through 75 separately for sMRI and fMRI. This resulted in 50 RSNs from the fMRI data and 45 ISNs from the sMRI data across both *modality-coupled* and *modality unique* sources. Next, we grouped these selected components into 7 subdomains based on their anatomical and functional properties, subcortical (SC), auditory (AU), sensorimotor (SM), visual (VI), default mode (DM), cognitive control (CC), cerebellum (CB). Figure 5 shows these components. Associated brain regions are also displayed in Table1.

Domains including SC, AU and CB have mostly highly linked components while other higher cognitive domains such as SM, VI, CC and DM have both highly linked and low linked components. This resulted in 44 modality coupled sources and 7 modality unique sources, 6 from fMRI and 1 from sMRI. This illustrates the power and flexibility of the proposed model, and suggests it is not artificially creating linkages, but rather encourages the linkage up until the data allows. The retained number of components was more for the fMRI case, this is expected as functional activity is expected to have more spatial variation due to its dynamic and distributed nature of neural activity.

Another notable outcome of CLiP-ICA is the relatively high number of retained good components, which resulted in fewer artifact components overall. CLiP-ICA identified 50 resting-state networks (RSNs) from a model order of 75, a significant achievement given that typically, we would need to analyze over 100 components to reach similar numbers. For instance, in previous studies, Allen et al. (2011) identified 28 RSNs out of 75 components, Rashid et al. (2018) found 38 RSNs out of 100 components, Agcaoglu et al. (2019) identified 51 RSNs out of 150 components, and Du et al. (2020) reported 53 RSNs from 100 components.

Although the retained variance was just slightly lower than what would be expected from a model order of 100 (since the principal components with smaller eigenvalues typically contribute very little additional variance), a greater number of high-quality RSNs were retained. This suggests that the inclusion of multimodal data and modeling components as joint multimodal sources helped to bring out the underlying brain networks. Supporting this interpretation, we observed that unlinked components (i.e., components numbered greater than 59) exhibited a higher ratio of artifact components. This suggests that the multimodal fusion not only improved the quality of the retained components but also effectively the specificity of the identified RSNs.

Another outcome is that CLiP-ICA retained more unlinked fMRI components compared to sMRI components. This is expected, as fMRI data typically exhibits higher variance due to temporal changes, resulting in a greater number of distinct sources. This finding also highlights the power and flexibility of the proposed CLiP-ICA method, which encourages linkages but does not force them where they do not naturally exist in the data.

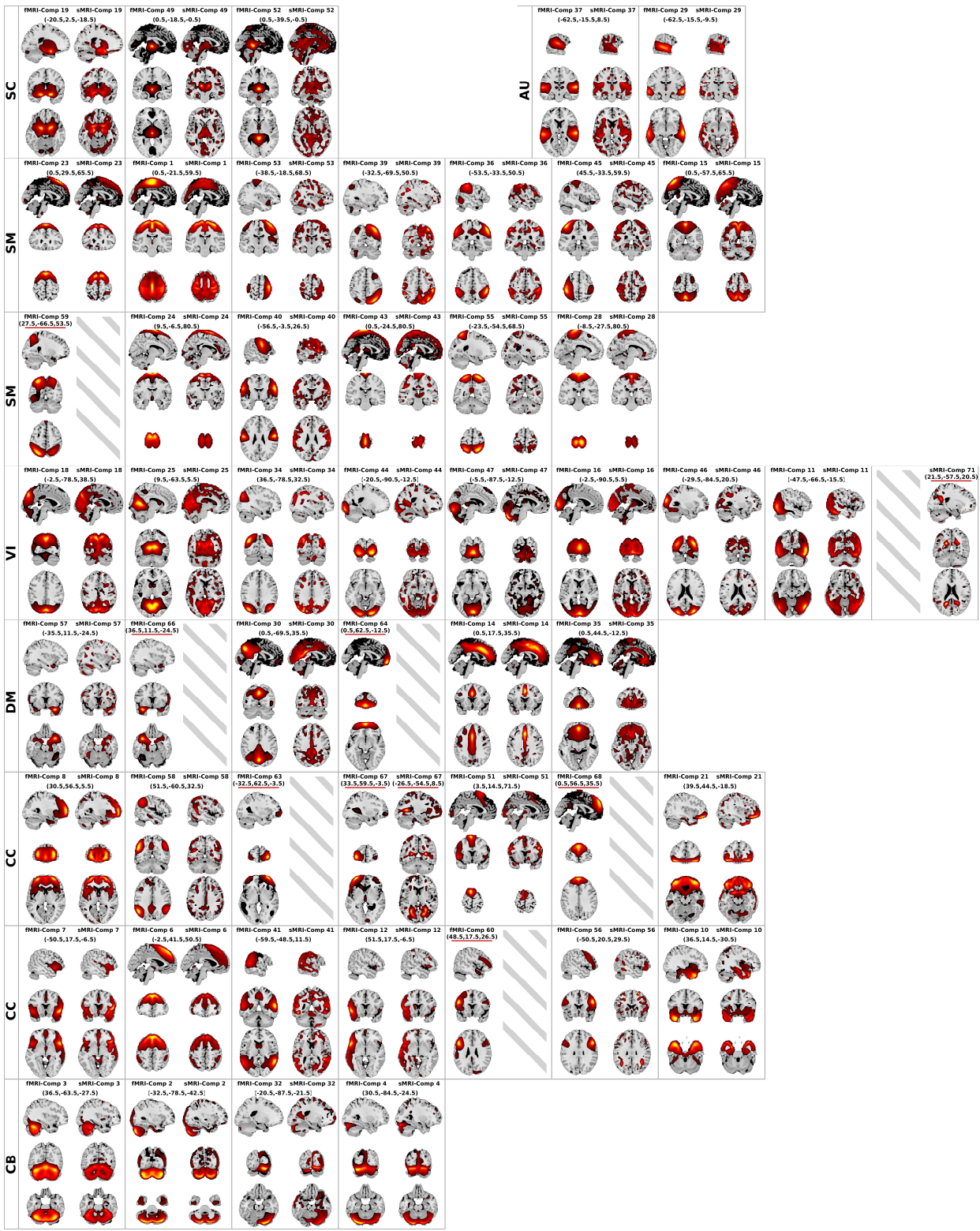

Figure 5: The selected RSNs and ISNs are grouped into seven subdomains and displayed as slices at their peak values, with fMRI and sMRI components placed side by side. Components numbered up to 59 are shown at the same coordinates for cross-modal comparison purposes, while components numbered 60 and above (i.e., modality-unique components) are displayed separately at their respective peaks in RSNs and ISNs. When a component is identified as a good component in only one modality, the corresponding component in the other modality is represented by a dashed outline.

Table 1: Brain regions associated with the RSNs and ISNs shown in Figure 4.

| Component Number | fMRI Coordinates | fMRI Label | sMRI Coordinate | sMRI Label | Linkage Status |
|---|---|---|---|---|---|
| SC - 19 | 21 2 -19 | R Uncus | 21 2 -19 | R Uncus | Linked |
| SC - 49 | 0 -19 -1 | L Thalamus | 0 -19 -1 | L Thalamus | Linked |
| SC - 52 | 0 -40 -1 | L Culmen | 0 -40 -1 | L Culmen | Linked |
| AU - 37 | 63 -16 8 | R Sup. Temp. Gyrus | 63 -16 8 | R Sup. Temp. Gyrus | Linked |
| AU - 29 | 63 -16 -10 | R Middle Temp. Gyrus | 63 -16 -10 | R Middle Temp. Gyrus | Linked |
| SM - 23 | 0 29 65 | Supp. Motor Area | 0 29 65 | Supp. Motor Area | Linked |
| SM - 1 | 0 -22 59 | L Medial Frontal Gyrus | 0 -22 59 | L Medial Frontal Gyrus | Linked |
| SM - 53 | 39 -19 68 | R Precentral Gyrus | 39 -19 68 | R Precentral Gyrus | Linked |
| SM - 39 | 33 -70 50 | R Sup. Parietal Lobule | 33 -70 50 | R Sup. Parietal Lobule | Linked |
| SM - 36 | 54 -34 50 | R Postcentral Gyrus | 54 -34 50 | R Postcentral Gyrus | Linked |
| SM - 45 | -45 -34 59 | L Postcentral Gyrus | -45 -34 59 | L Postcentral Gyrus | Linked |
| SM - 15 | 0 -58 65 | L Precuneus | 0 -58 65 | L Precuneus | Linked |
| SM - 59 | -27 -67 53 | L Sup. Parietal Lobule | | | fMRI only |
| SM - 24 | -9 -7 83 | Supp. Motor Area | -9 -7 83 | Supp. Motor Area | Linked |
| SM - 40 | 57 -4 26 | R Precentral Gyrus | 57 -4 26 | R Precentral Gyrus | Linked |
| SM - 43 | 0 -25 83 | Paracentral Lobule | 0 -25 83 | Paracentral Lobule | Linked |
| SM - 55 | 24 -55 68 | R Postcentral Gyrus | 24 -55 68 | R Postcentral Gyrus | Linked |
| SM - 28 | 9 -28 80 | R Paracentral Lobule | 9 -28 80 | R Paracentral Lobule | Linked |
| VI - 18 | 3 -79 38 | R Precuneus | 3 -79 38 | R Precuneus | Linked |
| VI - 25 | -9 -64 5 | L Lingual Gyrus | -9 -64 5 | L Lingual Gyrus | Linked |
| VI - 34 | -36 -79 32 | L Sup. Occipital Gyrus | -36 -79 32 | L Sup. Occipital Gyrus | Linked |
| VI - 44 | 21 -91 -13 | R Fusiform Gyrus | 21 -91 -13 | R Fusiform Gyrus | Linked |
| VI - 47 | 6 -88 -13 | R Lingual Gyrus | 6 -88 -13 | R Lingual Gyrus | Linked |
| VI - 16 | 3 -91 5 | R Cuneus | 3 -91 5 | R Cuneus | Linked |
| VI - 46 | 30 -85 20 | R Middle Occipital Gyrus | 30 -85 20 | R Middle Occipital Gyrus | Linked |
| VI - 11 | 48 -67 -16 | R Fusiform Gyrus | 48 -67 -16 | R Fusiform Gyrus | Linked |
| VI - 71 | | | -21 -57 21 | L Posterior Cingulate | sMRI only |
| DM - 57 | 36 11 -25 | R Sup. Temp. Gyrus | 36 11 -25 | R Sup. Temp. Gyrus | Linked |
| DM - 66 | -36 11 -25 | L Sup. Temp. Gyrus | | | fMRI only |
| DM - 30 | 0 -70 35 | L Precuneus | 0 -70 35 | L Precuneus | Linked |
| DM - 64 | 0 62 -13 | L Medial Frontal Gyrus | | | fMRI only |
| DM - 14 | 0 17 35 | L Cingulate Gyrus | 0 17 35 | L Cingulate Gyrus | Linked |
| DM - 35 | 0 44 -13 | L Medial Frontal Gyrus | 0 44 -13 | L Medial Frontal Gyrus | Linked |
| CC - 8 | -30 56 5 | L Middle Frontal Gyrus | -30 56 5 | L Middle Frontal Gyrus | Linked |
| CC - 58 | -51 -61 32 | L Angular Gyrus | -51 -61 32 | L Angular Gyrus | Linked |
| CC - 63 | 33 62 -4 | R Sup. Frontal Gyrus | | | fMRI only |
| CC - 67 | -33 59 -4 | L Sup. Frontal Gyrus | 27 -55 8 | R Parahippocampal Gyrus | Not Linked |
| CC - 51 | -3 14 71 | L Sup. Frontal Gyrus | -3 14 71 | L Sup. Frontal Gyrus | Linked |
| CC - 68 | 0 56 35 | L Sup. Frontal Gyrus | | | fMRI only |
| CC - 21 | -39 44 -19 | L Sup. Frontal Gyrus | -39 44 -19 | L Sup. Frontal Gyrus | Linked |
| CC - 7 | 51 17 -7 | R Sup. Temp. Gyrus | 51 17 -7 | R Sup. Temp. Gyrus | Linked |
| CC - 6 | 3 41 50 | R Sup. Frontal Gyrus | 3 41 50 | R Sup. Frontal Gyrus | Linked |
| CC - 41 | 60 -49 11 | R Sup. Temp. Gyrus | 60 -49 11 | R Sup. Temp. Gyrus | Linked |
| CC - 12 | -51 17 -7 | L Sup. Temp. Gyrus | -51 17 -7 | L Sup. Temp. Gyrus | Linked |
| CC - 60 | -48 17 26 | L Middle Frontal Gyrus | | | fMRI only |
| CC - 56 | 51 20 29 | R Middle Frontal Gyrus | 51 20 29 | R Middle Frontal Gyrus | Linked |
| CC - 10 | -36 14 -31 | L Sup. Temp. Gyrus | -36 14 -31 | L Sup. Temp. Gyrus | Linked |
| CB - 3 | -36 -64 -28 | L Tuber/Crus 1 | -36 -64 -28 | L Tuber/Crus 1 | Linked |
| CB - 2 | 33 -79 -43 | R Inf. Semi-Lunar Lobule | 33 -79 -43 | R Inf. Semi-Lunar Lobule | Linked |
| CB - 32 | 21 -88 -22 | R Declive | 21 -88 -22 | R Declive | Linked |
| CB - 4 | -30 -85 -25 | L Uvula/Crus 1 | -30 -85 -25 | L Uvula/Crus 1 | Linked |

C. *Functional Network Connectivity and sMRI loading*

Average group FNCs are presented in Figure 6. The analysis of FNC matrices reveals several notable patterns and differences across groups. The average FNC map demonstrates a highly modular structure, showing strong intra-network connectivity, particularly within the SM and VI networks. This modularity is consistent with previous studies (Cetin et al., 2016; Rashid et al., 2019; Du et al., 2020; Allen et al., 2018; Allen et al., 2011; Agcaoglu et al., 2019; Rashid et al.,

2018; Agcaoglu et al., 2018; Agcaoglu et al., 2020; Allen et al., 2014; Damaraju et al., 2014; Rashid et al., 2014; Sakoglu et al., 2010), further supporting the validity of the generated components. A strong negative connectivity pattern is observed between the DM and other networks such as SM and VI, providing further information regarding the brain's functional organization.

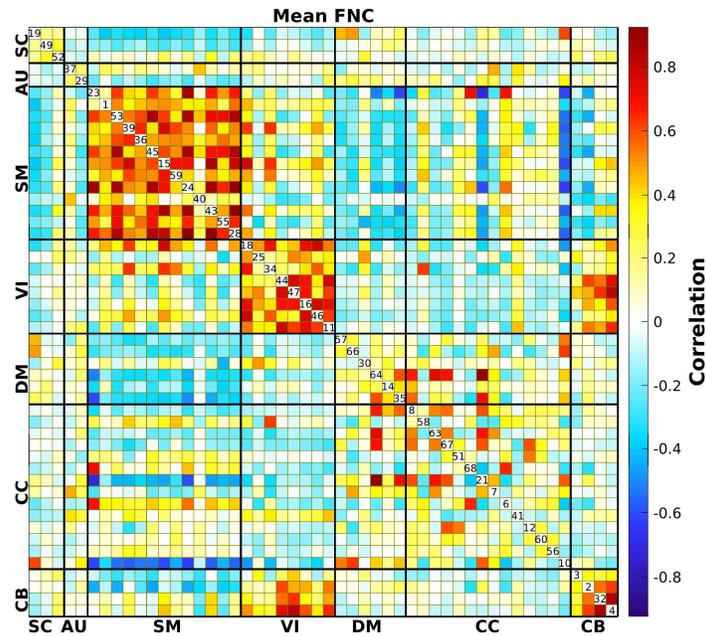

Figure 6: Average functional network connectivity. CLiP-ICA revealed a very modular FNC map, showing strong connectivity in SM, VI and CB networks. General patterns are consistent with the previous FNC analysis results showing strong connectivity in intra domain and negative connectivity between DM and other SM and VI.

We also tested group differences in FNC using two-sample t-test, we carried out t-test between CN and other sub-symptoms group, and corrected for multiple comparison using the false discovery rate (FDR) < 0.05. These results are presented in Figure 7 with symptom severity decreasing order; CN vs AD, LMCI, MCI, EMCI, SMC, and also all MCIs combined. In general, CN subjects exhibit higher connectivity, especially in SM and VI areas, compared to symptomatic groups. As the severity of symptoms decreases, the differences in connectivity also diminish, suggesting a gradient of functional decline. However, there are some exceptions to these, SM networks shows an increased in connectivity within the MCI group. Moreover, connectivity differences between SC #19 (uncus) and SM , and SM #1 (medial frontal gyrus) and other SM components also diminishes in MCI, but reappears in the AD stage. This could indicate a non-linear trajectory of connectivity changes, potentially reflecting compensatory mechanisms or heterogeneity in disease progression within the MCI stage. MCI has higher FNC in all intra-SM networks, though only few of the network pairs show FDR significant differences.

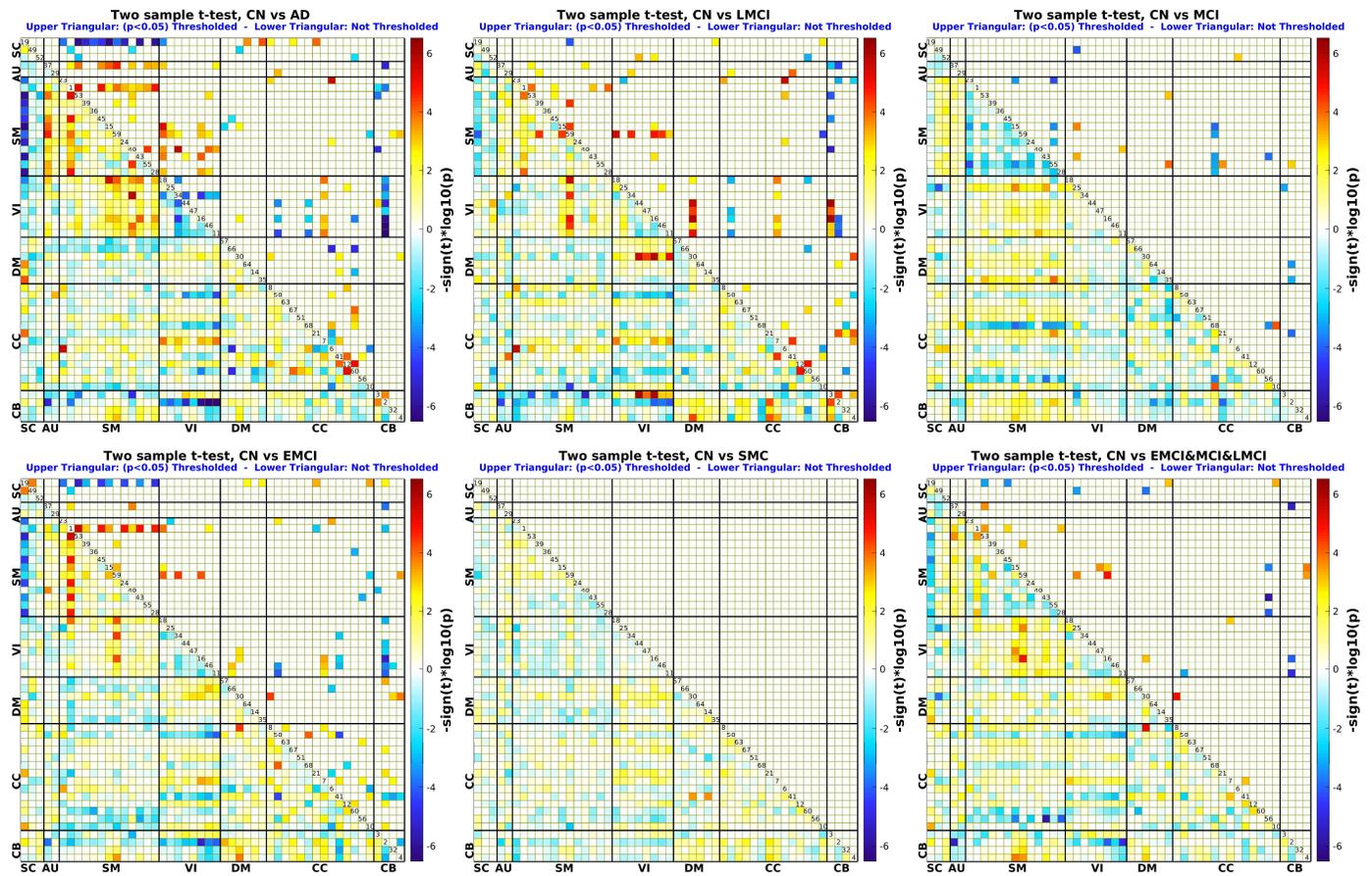

Figure 7: Results of two sample t-tests comparing connectivity differences between CN subjects and those with AD and its subgroups, organized by symptom severity: CN vs. AD, LMCI, MCI, EMCI, SMC and lastly all MCI subgroups. The upper triangular portion of the matrix displays thresholded $p$-values (FDR < 0.05) represented as -sign($T$)log10($p$), while the lower triangular portion presents unthresholded $p$-values in the same format; all figures have the same color bar range for convenient comparison. The analysis reveals significant connectivity differences between CN and all symptomatic groups except SMC. Generally, CN subjects exhibit higher connectivity, particularly in sensorimotor (SM) and visual (VI) areas. The differences in connectivity diminish as symptom severity decreases, although an unexpected pattern is observed in the MCI group; and SM connectivity with SM#1 and SC#19 in LMCI. In this group, the number of significant differences between pairs decreases, and there is a notable change in SM connectivity, indicating higher connectivity in MCI compared to other stages.

Further analyses comparing individuals at varying stages of cognitive decline with cognitively normal individuals reveal significant connectivity differences between MCI and AD; in the SM network, as well as between SM and subcortical (SC), visual (VI), and cerebellar (CC) regions. In general the pattern was similar for across all comparisons with controls. AD also appears to be more similar to the other cognitive decline groups. Results are presented in Figure 8. These findings highlight the complex and differential impact of AD on these networks across subgroups, suggesting that the progression of the disease may uniquely affect these specific areas depending on the stage and severity of cognitive decline. As an example of the complementary information contained in the different modalities, the loadings of SM #1 show no significant group differences in the loadings between any of the other cognitive groups. However, FNC of SM #1 with other SM networks showed a prominent pattern of significant group differences between CN vs

EMCI, which disappears in MCI and LMCI, and reappears in AD. SM #1 is the most strongly linked component between sMRI and fMRI modalities. This suggest that while the structural integrity of SM #1 may not be impacted significantly by cognitive decline, its connectivity significantly changes with other SM networks (which do show changes in sMRI loadings). Moreover, SM #23 and SM #24 show a significant increase in loading in AD, compared to CNs. SM #23 is located adjacent and anterior to SM#1. This illustrates the benefits of using the multimodal approach and complementary information provided by sMRI and fMRI.

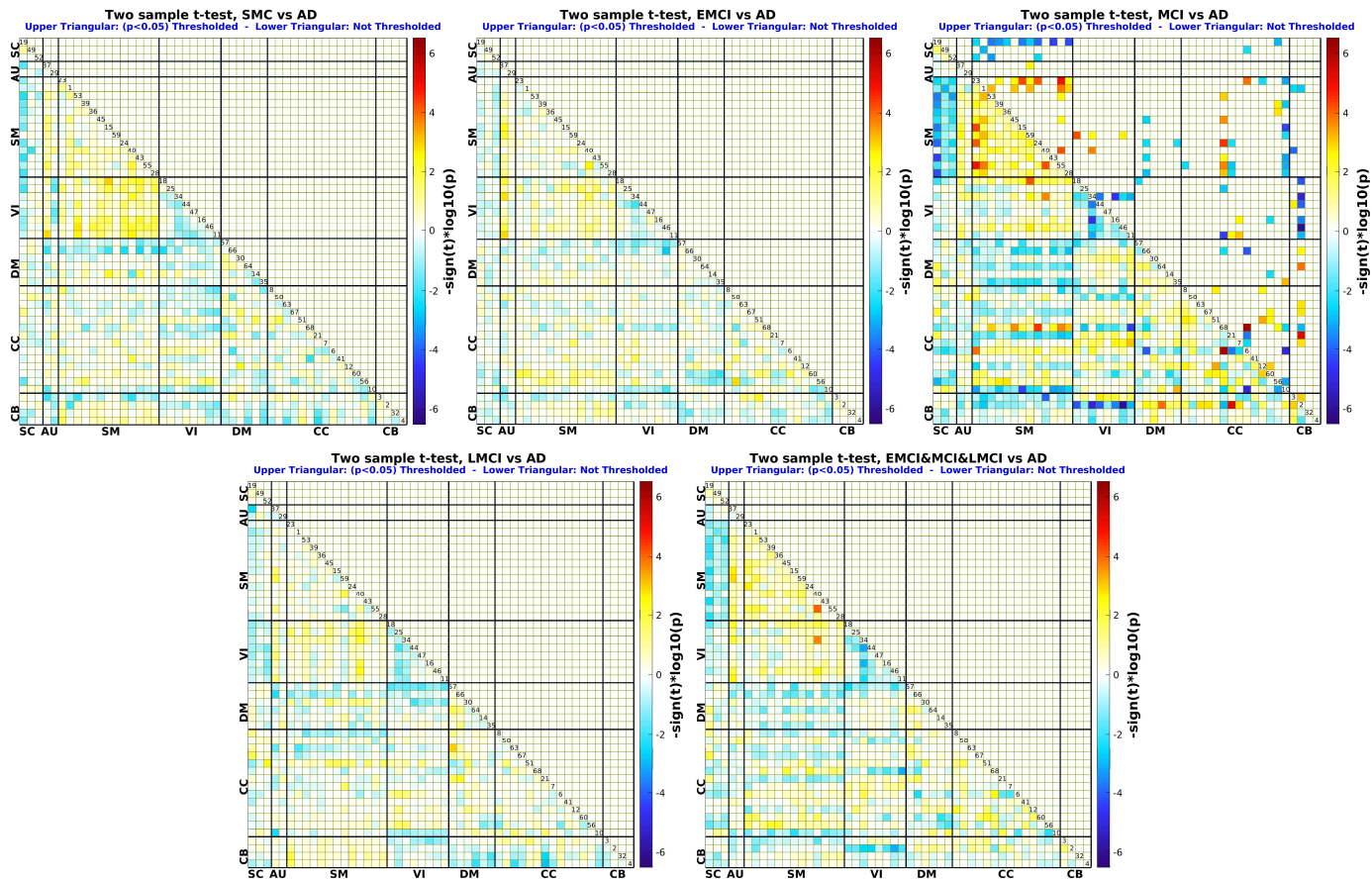

Figure 8: Results of two sample t-tests comparing connectivity differences between AD subjects and those with its subgroups, organized by symptom severity: AD vs. LMCI, MCI, EMCI, SMC and lastly all MCI subgroups. The upper triangular portion of the matrix displays thresholded $p$-values (FDR < 0.05) represented as $-sign(T)log10(p)$, while the lower triangular portion presents unthresholded $p$-values in the same format; all figures have the same color bar range for convenient comparison. The analysis reveals significant connectivity differences between AD and MCI, particularly in SM and SM with SC, VI and CC.

sMRI loading differences between CN individuals and those with varying degrees of cognitive impairment are displayed in Figure 9. Significant differences in sMRI loadings are observed across most subgroups, except for the SMC group. CN subjects generally show higher sMRI loadings, which reflect greater preservation of brain structure and lower levels of atrophy compared to other groups. This suggests that CN individuals maintain more structural integrity in brain regions affected by cognitive decline. In contrast, subjects in subgroups such as AD, LMCI, MCI, and EMCI display lower sMRI loadings, indicating reduced structural integrity in specific brain regions.

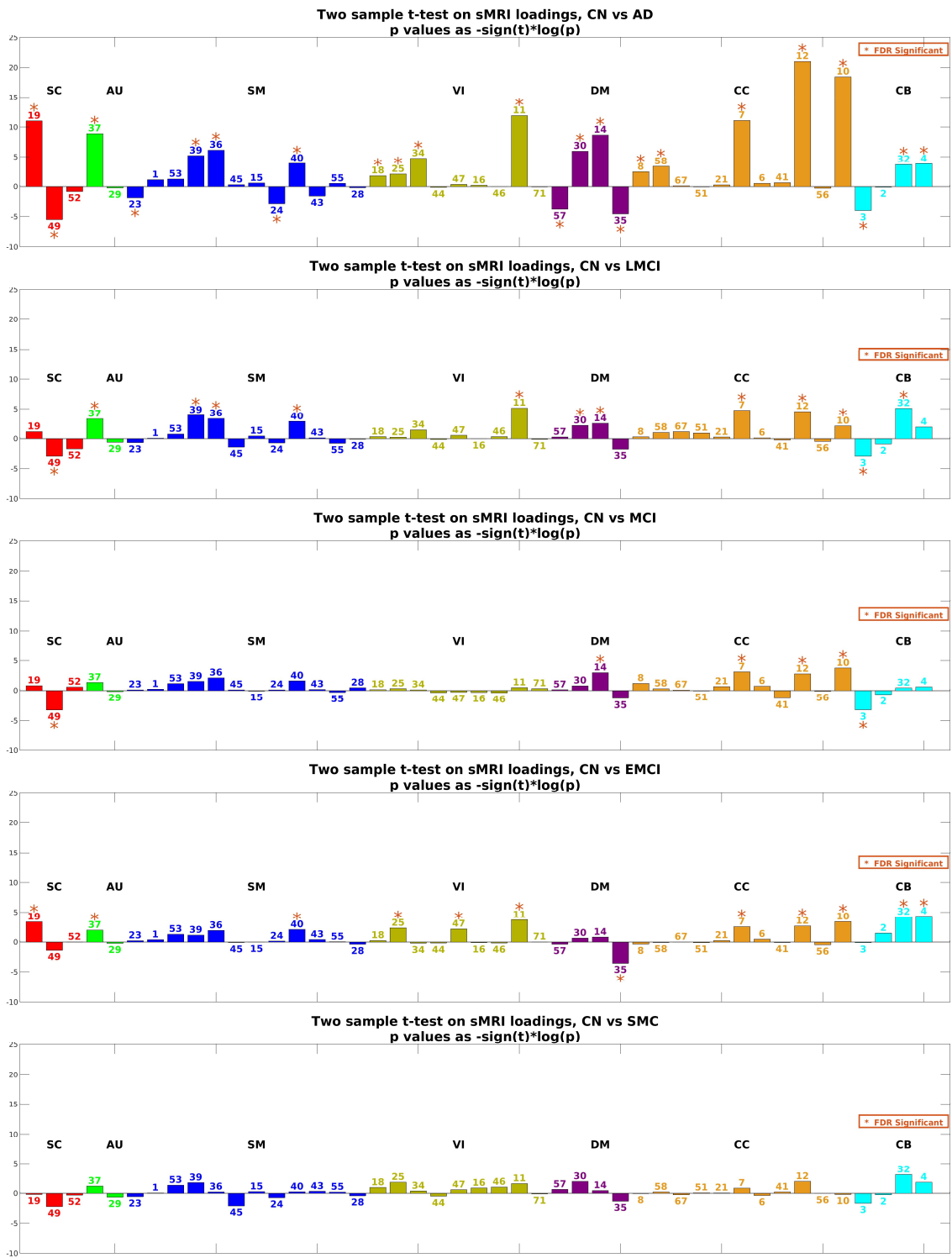

Figure 9: Results of two sample t-tests comparing sMRI loading differences between CN subjects and those with AD and its subgroups, organized by symptom severity: CN vs. AD, LMCI, MCI, EMCI and SMC. *P*-values represented as $-sign(T)log10(p)$, and those who showed significant differences (FDR < 0.05) are marked with '*'. Results show significant differences in all subgroups except SMC vs CN. We observe significant group differences between CN and AD in almost all domains, mostly showing higher loadings in CN.

The group mean SNC is shown in Figure 10 on the left where the upper triangular shows the thresholded (p<0.05) correlation values and the lower triangular shows the unthresholded correlation values, and group differences between CN and AD subjects on the right where the upper triangular shows the thresholded correlation values (p<0.05 assessed with Fisher's z-test) and lower triangular shows the unthresholded correlation values. SNC is a metric designed to capture patterns of covarying structural atrophy among structural brain networks, by calculating the Pearson correlation of sMRI loading parameters, that are generated using structural and functional coupling by CLiP-ICA. It provides a unique perspective by analyzing how structural variations in the brain align with functional networks. The left side of the figure shows a modular structure, with high levels of covariation observed particularly in the sensorimotor (SM) and visual (VI) networks, indicating strong structural co-existence within these regions. This is consistent with FNC results, which showed strong connectivity in SM and VI networks. In contrast, subcortical (SC) networks exhibit low covariation, suggesting lesser structural coupling. The right-side displays group differences in SNC as the subtraction between mean values for CN and AD groups. This comparison reveals a complex pattern of differences, highlighting areas where structural integrity is more preserved in CN subjects compared to those with AD. The observed variations in SNC reflect the progressive structural changes that accompany cognitive decline in AD, making it a promising tool for understanding how structural degeneration correlates with functional deficits across different brain regions.

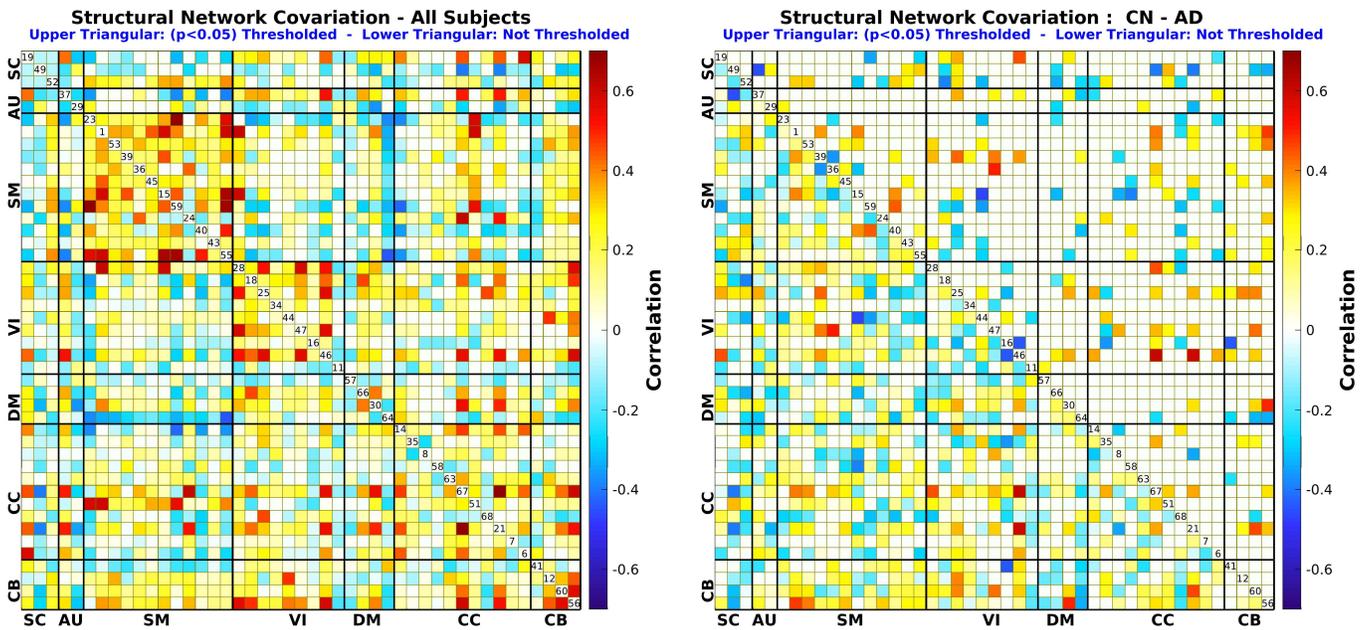

Figure 10: Structural network covariation (SNC) for all subjects is displayed on the left: upper triangular shows those statistically greater than 0 (p<0.05), lower triangular shows unthresholded values. We observed a modular structure with high covariation in the SM and VI networks, while the SC networks exhibited very low covariation. Group differences in SNC are shown on the right as the subtraction between mean CN and mean AD values; upper triangular shows the pairs that show statistical differences (p<0.05 assessed with Fisher's z-test) and lower triangular shows unthresholded values revealing a complex pattern of differences.

## IV. DISCUSSION

In this paper, we introduce and implement a novel method for multimodal fusion and provided applications on both simulation and real dataset to illustrate its capabilities for coupling structural and functional neuroimaging data. The CLiP-ICA method offers several advantages in the fusion of sMRI and fMRI imaging data. Unlike traditional fusion methods, which often reduce 4D fMRI data to 3D features, CLiP-ICA fully integrates the rich temporal dynamics of fMRI, preserving critical information about time-varying brain activity. By utilizing a copula-based approach, CLiP-ICA flexibly links spatial components from different modalities, capturing both shared and unique aspects of brain networks.

The findings of this study demonstrate the effectiveness of the proposed CLiP-ICA in coupling structural and functional MRI data, providing insights into the complex interactions between brain structure and function across different stages of AD. Our approach was able to uncover both linked and modality-unique components, offering a rich understanding of the links between brain structure and brain network dynamics which are not accessible in traditional, separately applied, ICA methods. Previous studies (Luo, Sui, Abrol, Chen, et al., 2020; Kotoski et al., 2024; Segall et al., 2012; Luo, Sui, Abrol, Lin, et al., 2020) reported the intrinsic networks in resting fMRI, also observed in sMRI data. Our approach enables the simultaneous estimation of these networks, allowing for a more effective analysis while preserving the temporally rich data from 4D fMRI. In contrast, other methods often reduce 4D fMRI data to 3D features, failing to fully leverage the complete temporal information during data fusion, especially in cases of mismatched dimensionality

CLiP-ICA identified a higher number of meaningful components in both the functional and structural domains, even with a relatively low model order. The ability to couple structural and functional information during network estimation proved advantageous, offering benefits in addressing the long-standing challenge of determining the optimal model order for such analyses. While CLiP-ICA retained a high number of RSNs and ISNs, it also effectively distinguished between modality-unique sources, capturing in the components above 59 with low sMRI/fMRI coupling. This ability to capture both shared and distinct neural characteristics underscores the flexibility of our approach in handling complex multimodal data.

CLiP-ICA also proved capable of accurately retrieving both highly linked and unlinked components, even when the assigned dependency values did not perfectly match the true cross-modality relationships. This adaptability highlights the robustness of the method, ensuring that it captures the intrinsic patterns of the data rather than imposing artificially high correlations. This is particularly important in studies of complex brain conditions like AD, where the nature of structural and functional coupling may vary substantially across different stages of the disease.

Furthermore, our analysis of FNC revealed a highly modular structure in the SM and VI networks, with notable negative connectivity between the DM network and these regions. Such modularity has been reported in previous studies (Rashid et al., 2019; Allen et al., 2011; Agcaoglu et al., 2019; Rashid et al., 2018; Agcaoglu et al., 2018; Sakoglu et al., 2010; Espinoza et al., 2019; López-Vicente et al., 2021), further validating the robustness of the CLiP-ICA approach. The observed connectivity differences between CN and symptomatic groups suggest that functional decline follows a gradient, with connectivity reductions becoming more pronounced as the severity of cognitive impairment increases.

Interestingly, the MCI group demonstrated a transient increase in connectivity within the SM network, suggesting possible compensatory mechanisms or variability in disease progression at this stage. Notably, our findings suggest that the progression of AD involves a non-linear pattern of connectivity changes. For example, while connectivity differences between SM and SC networks were reduced in the MCI stage, they reappeared in advanced stages of AD, pointing to a possible heterogeneous trajectory of neurodegeneration or adaptive processes that alter the course of the disease.

Overall, this study illustrates the potential of CLiP-ICA as a powerful tool for multimodal fusion in neuroimaging research. By integrating structural and functional MRI data, the method enables a more comprehensive analysis of the intricate changes that occur during the progression of AD. Future studies could build on these findings by applying CLiP-ICA to larger, more diverse datasets, and by exploring its application in longitudinal studies to better track the dynamics of brain changes and assess the impact of therapeutic interventions over time.

Some limitations should be considered when interpreting the results of this study. First, the dataset used includes unbalanced group sizes, with a higher number of CN subjects compared to those in other dementia subgroups. This imbalance could potentially bias the results, as differences between groups may be influenced by the disproportionate sample sizes. Future studies should aim to include a more balanced representation of each clinical group to ensure more robust comparisons. Additionally, we used a Gaussian copula to link the components, which assumes uniform contribution across the entire distribution for a given source-pairs. However, brain networks are often quite spares, with the most important values occurring at the tails of the logistic distribution. By using a Gaussian copula, we might have underestimated these tail-dependent interactions. Exploring alternative copula models that place more emphasis on the tails of the distribution, such as FGM copula, could potentially provide a more accurate representation of the complex dependencies between structural and functional networks.

## V. Conclusion

In summary, we introduced the CLiP-ICA method, designed to maximize the use of the entire 4D fMRI dataset for the integration of fMRI and sMRI by linking network spatial patterns through a copula-based approach. The performance of the CLiP-ICA method was evaluated using both simulation studies and real-world data, with comparisons to results obtained from running separate ICAs on each modality. Our analysis demonstrated that CLiP-ICA successfully captured FNC and loading parameters, revealing significant group differences among clinical groups, including cognitively normal individuals and those with varying levels of cognitive impairment. When applied to real data, our approach uncovered several intriguing findings, such as the presence of highly linked components and modality-unique components. Additionally, we observed distinct group differences across a variety of FNCs, further emphasizing the method's ability to detect subtle variations in brain connectivity associated with different stages of cognitive decline. Overall, CLiP-ICA provides a powerful tool for fusing structural and functional MRI data, offering new insights into the complex interplay between brain structure and function. This method not only enhances our understanding of neurodegenerative processes but also holds potential for broader applications in other neurological conditions, making it a promising approach for future research in multimodal neuroimaging.

**ACKNOWLEDGMENTS**

This work was supported by NIH RF1AG063153 and NSF 2112455.

# Supplementary Materials

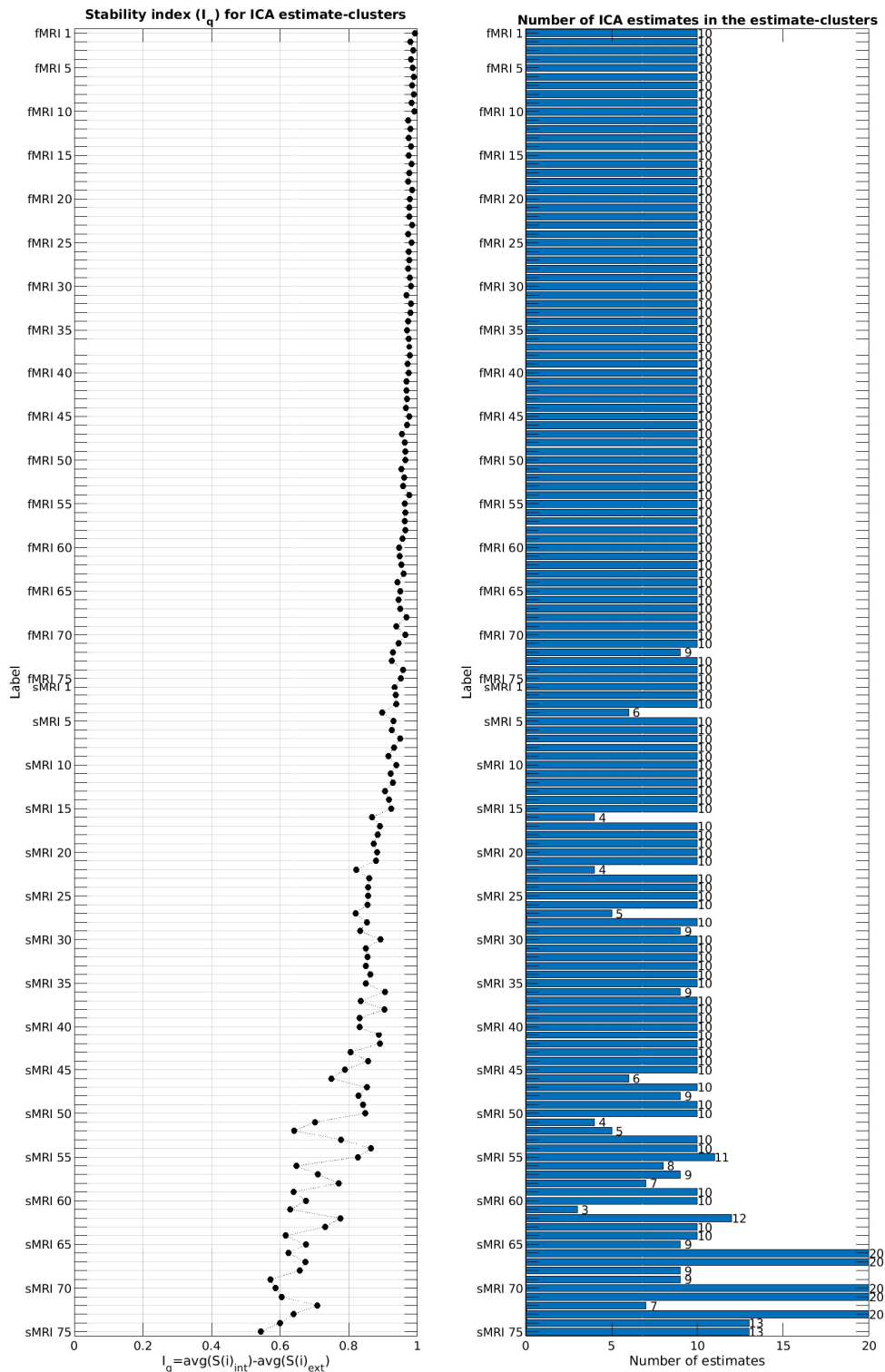

Sup. Figure 1: The stability index of the CLiP-ICA components was derived from ICASSO estimation over the 10 different runs of CLiP-ICA with random initialization. Overall, the results indicate that the component stability is satisfactory. Specifically, fMRI components demonstrate greater stability compared to sMRI components, and linked components exhibit higher stability than their unlinked counterparts.

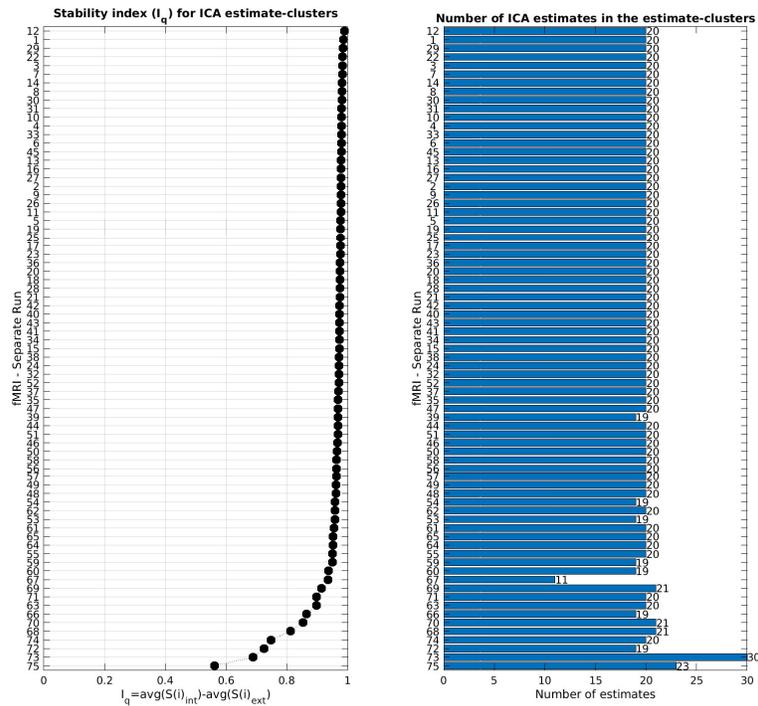

Sup. Figure 2: The stability index of the fMRI components with separate runs, was derived from ICASSO estimation over the 20 different runs of infomax with random initialization. Overall, the results are consistent with those in CLiP-ICA, showing strong stability in majority of components. The stability index in separate cases is smaller than those in the CLiP-ICA runs. This shows the benefit of the multimodal coupling in spatial domain.

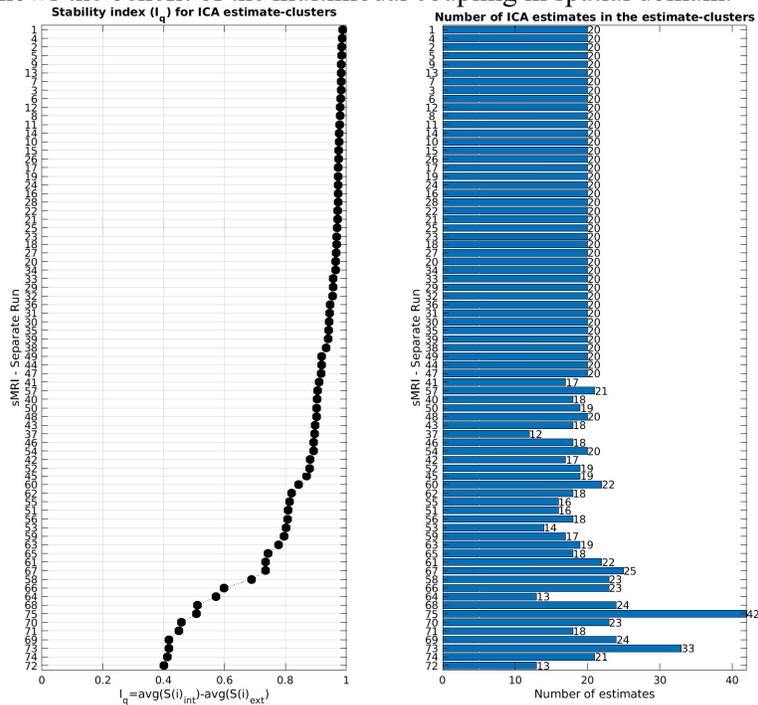

Sup. Figure 3: The stability index of the sMRI components with separate runs, was derived from ICASSO estimation over the 20 different runs of infomax with random initialization. Overall, the results are consistent with those in CLiP-ICA. Though sMRI showed less robustness compared to the fMRI, majority of the components are still very stable. The stability index in separate cases is smaller than those in the CLiP-ICA runs. This shows the benefit of the multimodal coupling in spatial domain.

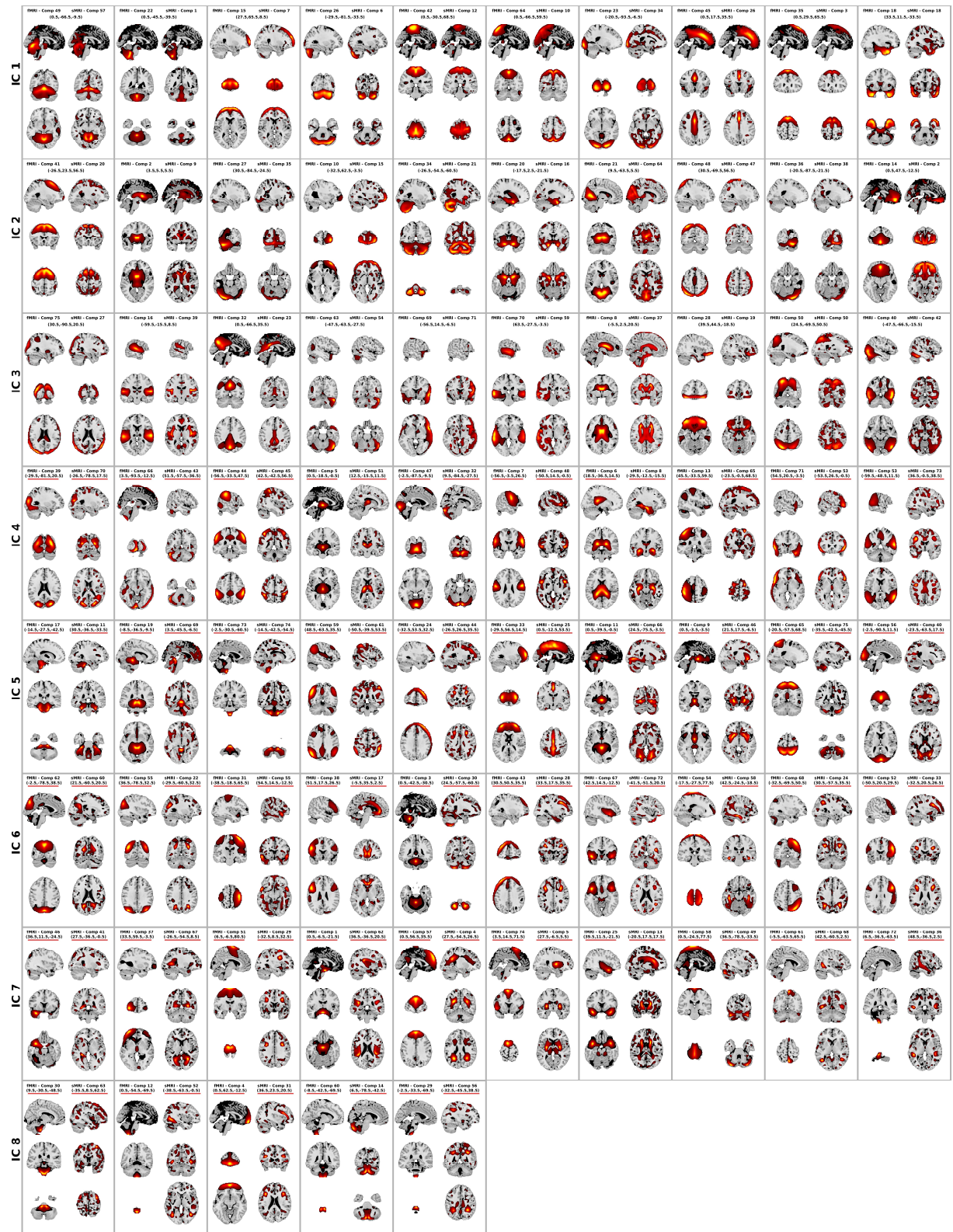

Sup. Figure 4: The fMRI and sMRI components generated running gICA separately and then matching components post-hoc; are displayed side by side at slices at their peak values. The components are ordered showing the highest cross correlation to lowest cross correlation. The first 30 components with the highest cross correlation which have higher Pearson correlation value than 0.25 are displayed at the common slices, while the remainder ones are displayed as at their individual fMRI and sMRI peak location.